\documentclass[twoside,11pt]{article}

\usepackage{blindtext}

\usepackage[abbrvbib]{jmlr2e}
\usepackage{jmlr2e}
\usepackage{listings}
\usepackage{xcolor}
\usepackage{xspace}
\usepackage{makecell}
\usepackage{booktabs}
\usepackage{graphicx}
\usepackage{lipsum}
\usepackage{subcaption}
\usepackage{multirow}

\definecolor{dkgreen}{rgb}{0,0.6,0}
\definecolor{mauve}{rgb}{0.58,0,0.82}
\usepackage{lastpage}

\ShortHeadings{SocialED: A Python Library for Social Event Detection}{Zhang, Yu, Li, Peng, and Yu}
\firstpageno{1}

\begin{document}

\title{SocialED: A Python Library for Social Event Detection}


\author{\name Kun Zhang$^1$                                 \email zhangkun23@buaa.edu.cn       \\
        \name Xiaoyan Yu$^2$                                \email xiaoyan.yu@bit.edu.cn        \\
        \name Pu Li$^3$                                     \email lip@stu.kust.edu.cn          \\
        \name Hao Peng$^1$\thanks{Corresponding author.}    \email penghao@buaa.edu.cn          \\
        \name Philip S. Yu$^4$                              \email psyu@uic.edu                 \\
        \addr $^1$Beihang University,
        \addr $^2$Beijing Institute of Technology,
        \addr $^3$Kunming University of Science and Technology,
        \addr $^4$University of Illinois at Chicago
       }

\editor{My editor}

\maketitle

\begin{abstract}

    SocialED is a comprehensive, open-source Python library designed to support social event detection (SED) tasks, integrating 19 detection algorithms and 14 diverse datasets. 
    It provides a unified API with detailed documentation, offering researchers and practitioners a complete solution for event detection in social media. 
    The library is designed with modularity in mind, allowing users to easily adapt and extend components for various use cases.
    SocialED supports a wide range of preprocessing techniques, such as graph construction and tokenization, and includes standardized interfaces for training models and making predictions.
    By integrating popular deep learning frameworks, SocialED ensures high efficiency and scalability across both CPU and GPU environments. 
    The library is built adhering to high code quality standards, including unit testing, continuous integration, and code coverage, ensuring that SocialED delivers robust, maintainable software. 
    SocialED is publicly available at \url{https://github.com/RingBDStack/SocialED} and can be installed via PyPI.
    
\end{abstract}


\section{Introduction}

    Social event detection \citep{hasan2018survey} is the task of identifying noteworthy events from social media platforms, which is typically approached as a classification \citep{ren2022evidential} or clustering \citep{cao2021knowledge,ren2022known} task. 
    SED plays a crucial role in a wide range of downstream applications, such as crisis management \citep{abavisani2020multimodal}, public opinion monitoring \citep{karamouzas2022public}, fake news detection \citep{liu2020fned}, etc. 
    Its broad applicability has attracted significant attention from both the academic research community and industry practitioners \citep{peng2022reinforced}. 

    To cope with the increasing demand for SED, several powerful algorithms and models have been applied and proposed over the years, ranging from general methods like LDA \citep{blei2003latent}, 
    BiLSTM \citep{graves2005framewise}, Word2Vec \citep{miklov2013efficient}, 
    and BERT \citep{devlin2018bert}, to specialized approaches such as KPGNN \citep{cao2021knowledge}, QSGNN \citep{ren2022known}, FinEvent \citep{peng2022reinforced}, and HISEvent \citep{cao2024hierarchical}. 
    Despite these advancements, deploying these methods or conducting a comprehensive evaluation remains challenging, as a unified evaluation framework and a comprehensive toolkit that can integrate these diverse approaches are still absent.
    In contrast, other tasks, such as Graph Outlier Detection \citep{liu2024pygod} and Scalable Outlier Detection \citep{zhao2019pyod}, have well-established libraries and toolkits that provide standardized evaluation metrics and modular implementations. 
    
    To fill this gap, we present \textbf{SocialED}, an open-source Python library designed to facilitate the development and evaluation of \textbf{Social} \textbf{E}vent \textbf{D}etection algorithms.
    SocialED up-to-now integrates 19 detection algorithms and supports 14 widely-used datasets, as shown in Table \ref{tab:algorithms}, with plans for continuous updates to include emerging datasets and cutting-edge methods.
    Specifically, SocialED offers a comprehensive range of features to meet the diverse needs of social event detection research.
    \textbf{Firstly}, SocialED integrates a wide range of SED datasets and algorithms, offering users versatile tools for their research.
    \textbf{Secondly}, SocialED implements these algorithms with a unified API, allowing seamless data preparation and integration across all models.
    \textbf{Thirdly}, SocialED provides modular, customizable components for each algorithm, allowing users to tailor models to their specific needs.
    \textbf{Moreover}, SocialED offers a set of utility functions designed to simplify the construction of social event detection workflows.
    \textbf{Finally}, SocialED provides comprehensive API documentation, examples, unit tests, and maintainability features, ensuring robust and reliable code.

\begin{table}[!t]
\small
\begin{subtable}[t]{0.6\textwidth}
\setlength{\tabcolsep}{3pt}
\renewcommand{\arraystretch}{1.11}
\begin{tabular}{l|c|c}
     \toprule
    \textbf{Algorithms}                                         & \textbf{Backbone}     & \textbf{Supervision}  \\
    \midrule
    \textbf{LDA}        \tiny \citep{blei2003latent}        & Topic                 & Unsupervised          \\
    \textbf{BiLSTM}     \tiny \citetext{Graves et al., \citeyear{graves2005framewise}}   & DL                    & Supervised            \\
    \textbf{Word2Vec}   \tiny \citep{miklov2013efficient}   & WE                    & Unsupervised          \\
    \textbf{GloVe}      \tiny \citep{pennington2014glove}   & WE                    & Unsupervised          \\
    \textbf{WMD}        \tiny \citep{kusner2015word}        & Similarity            & Unsupervised          \\
    \textbf{BERT}       \tiny \citep{devlin2018bert}        & PLMs                  & Unsupervised          \\
    \textbf{SBERT}      \tiny \citetext{Reimers et al., \citeyear{reimers2019sentence}}   & PLMs                  & Unsupervised          \\
    \textbf{EventX}     \tiny \citep{liu2020story}          & CD                    & Unsupervised          \\
    \textbf{CLKD}       \tiny \citep{ren2024toward}         & GNNs                  & Supervised            \\
    \textbf{KPGNN}      \tiny \citep{cao2021knowledge}      & GNNs                  & Supervised            \\
    \textbf{FinEvent}   \tiny \citep{peng2022reinforced}    & GNNs                  & Supervised            \\
    \textbf{QSGNN}      \tiny \citep{ren2022known}          & GNNs                  & Supervised            \\
    \textbf{ETGNN}      \tiny \citep{ren2022evidential}     & GNNs                  & Supervised          \\
    \textbf{UCLSED}     \tiny \citep{ren2023uncertainty}    & GNNs                  & Supervised          \\
    \textbf{HCRC}       \tiny  \citep{guo2024unsupervised}  & GNNs                  & Unsupervised          \\
    \textbf{RPLMSED}    \tiny \citep{li2024relational}      & PLMs                  & Supervised            \\
    \textbf{HISEvent}   \tiny \citep{cao2024hierarchical}   & CD                    & Unsupervised          \\
    \textbf{ADPSEM}     \tiny \citep{yang2024adaptive}      & CD                    & Unsupervised          \\
    \textbf{HyperSED}   \tiny \citep{yu2025towards}         & \makecell{CD}         & Unsupervised          \\
    \bottomrule
\end{tabular}
\end{subtable}
\hfill
\begin{subtable}[t]{0.38\textwidth}
\setlength{\tabcolsep}{2pt}
\renewcommand{\arraystretch}{0.7}
\begin{tabular}{l|c|r}
    \toprule
    \textbf{Datasets}        & \#\textbf{Events} & \#\textbf{Texts}   \\
    \midrule
    \textbf{Event2012}              & \multirow{2}{*}{503}    & \multirow{2}{*}{68,841}  \\
    \tiny \raisebox{0pt}[0pt][0pt]{\citep{mcminn2013building}}&&\\
    \textbf{Event2018}               & \multirow{2}{*}{257}    & \multirow{2}{*}{64,516}  \\
    \tiny \raisebox{0pt}[0pt][0pt]{\citep{mazoyer2020french}} &&\\
    \textbf{ArabicTwitter}     & \multirow{2}{*}{7}      & \multirow{2}{*}{9,070}   \\
    \tiny \raisebox{0pt}[0pt][0pt]{\citep{alharbi2021kawarith}} &&\\
    \textbf{MAVEN}          & \multirow{2}{*}{164}    & \multirow{2}{*}{10,242}  \\
    \tiny \raisebox{0pt}[0pt][0pt]{\citep{wang2020maven}} &&\\
    \textbf{CrisisLexT26}     & \multirow{2}{*}{26}     & \multirow{2}{*}{27,933}  \\
    \tiny \raisebox{0pt}[0pt][0pt]{\citep{olteanu2015expect}} &&\\
    \textbf{CrisisLexT6}     & \multirow{2}{*}{6 }     & \multirow{2}{*}{60,082} \\
    \tiny \raisebox{0pt}[0pt][0pt]{\citep{olteanu2014crisislex}} &&\\
    \textbf{CrisisMMD}            & \multirow{2}{*}{7}      & \multirow{2}{*}{18,082} \\
    \tiny \raisebox{0pt}[0pt][0pt]{\citep{alam2018crisismmd}} &&\\
    \textbf{CrisisNLP}      & \multirow{2}{*}{11}     & \multirow{2}{*}{25,976}  \\
    \tiny \raisebox{0pt}[0pt][0pt]{\citep{alam2021crisisbench}} &&\\
    \textbf{HumAID}               & \multirow{2}{*}{19}     & \multirow{2}{*}{76,484}  \\
    \tiny \raisebox{0pt}[0pt][0pt]{\citep{alam2021humaid}} &&\\
    \textbf{MixData}          & \multirow{3}{*}{5}      & \multirow{3}{*}{78,489}  \\
    \tiny \raisebox{0pt}[0pt][0pt]{\citep{imran2013extracting}} &&\\
    \tiny \raisebox{0pt}[0pt][0pt]{\citep{alam2018twitter,alam2018graph}} &&\\
    \textbf{KBP}                & \multirow{2}{*}{100}    & \multirow{2}{*}{85,569}   \\
    \tiny \raisebox{0pt}[0pt][0pt]{\citep{deng2020meta}} && \\
    \textbf{Event2012\_100}     & \multirow{2}{*}{100}    & \multirow{2}{*}{15,019}  \\
    \tiny \raisebox{0pt}[0pt][0pt]{\citep{mcminn2013building}} && \\
    \textbf{Event2018\_100}    & \multirow{2}{*}{100}    & \multirow{2}{*}{19,944}   \\
    \tiny \raisebox{0pt}[0pt][0pt]{\citep{mazoyer2020french}} && \\
    \textbf{Arabic\_7}          & \multirow{2}{*}{7}      & \multirow{2}{*}{3,022}   \\
    \tiny \raisebox{0pt}[0pt][0pt]{\citep{alharbi2021kawarith}} && \\
    \bottomrule
\end{tabular}
\end{subtable}
\caption{Social event detection algorithms and datasets covered by SocialED, where DL, WE, and CD refer to Deep Learning, Word Embedding, and Community Detection, respectively. \label{tab:algorithms}}
\vspace{-1cm}
\end{table}

\begin{figure*}[htbp]
\centering
\resizebox{0.85\textwidth}{!}{
\includegraphics[width=\linewidth]{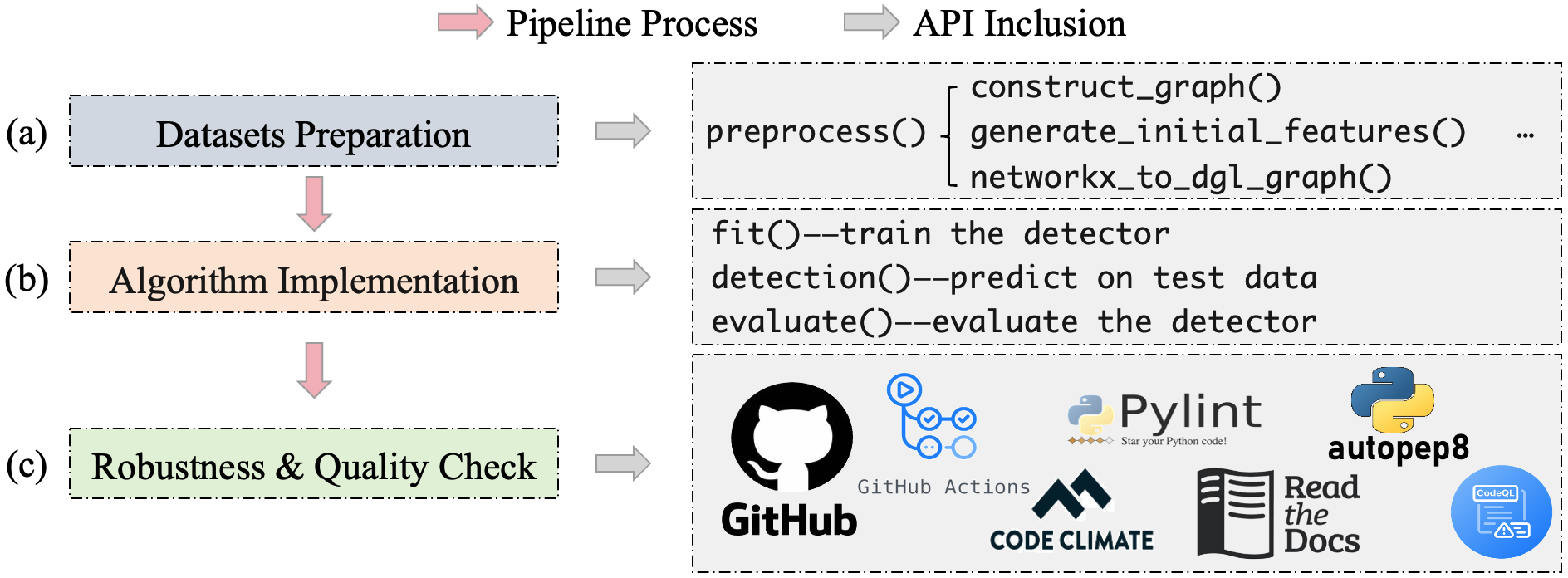}
}
\caption{Demonstration of SocialED's unified API design. }
\vspace{-0.5cm}
\label{fig:vis}
\end{figure*}

\section{Library Design and Implementation}

    \paragraph{Dependencies and Technology Stack}
    SocialED is compatible with Python 3.8 and above, and leverages well-established deep learning frameworks like \texttt{PyTorch} \citep{paszke2019pytorch} and \texttt{Hugging Face Transformers} \citep{wolf2019huggingface} for efficient model training and inference, supporting both CPU and GPU environments.
    In addition to these core frameworks, SocialED also integrates \texttt{NumPy}, \texttt{SciPy}, and \texttt{scikit-learn} for data manipulation, numerical operations, and machine learning tasks, ensuring versatility and performance across a range of workflows.

    \paragraph{Unified API Design}
    Inspired by the API designs of established frameworks \citep{sklearn_api,liu2024pygod}, we developed a unified API for all detection algorithms in SocialED: 
        (1) \texttt{preprocess} provides a flexible framework for handling various preprocessing tasks, such as graph construction and tokenization (Figure \ref{fig:vis}(a)). 
        Since different SED algorithms require distinct preprocessing steps, our modular design accommodates each model's specific needs while maintaining a unified interface.
        (2) \texttt{fit} trains the detection algorithms on the preprocessed data, adjusting model parameters and generating necessary statistics for predictions.
        (3) \texttt{detection} uses the trained model to identify events from the input data, returning the detected events.
    
    An example of the API usage is shown in Code Demo \ref{lst:kpgnn_event2012}.

\begin{minipage}{0.95\linewidth}
\renewcommand{\lstlistingname}{Code Demo}
\begin{lstlisting}[caption={Using KPGNN on the MAVEN dataset with SocialED API.},captionpos=b, label={lst:kpgnn_event2012}, numbers=left, xleftmargin=0.5em,frame=single,framexleftmargin=1.3em]
    from SocialED.dataset import MAVEN    
    dataset = MAVEN().load_data()           # Load "MAVEN" dataset
    from SocialED.detector import KPGNN     # Import KPGNN model
    args = args_define().args               # Get training arguments
    kpgnn = KPGNN(args, dataset)            # Initialize KPGNN model
    kpgnn.preprocess()                      # Preprocess data
    kpgnn.fit()                             # Train the model
    pres, trus = kpgnn.detection()          # Detect events
    kpgnn.evaluate(pres, trus)              # Evaluate detection results

\end{lstlisting}
\end{minipage}

    \paragraph{Modular Design and Utility Functions}
    SocialED is built with a modular design to improve reusability and reduce redundancy. 
    It organizes social event detection into distinct modules: \texttt{preprocessing}, \texttt{modeling}, and \texttt{evaluation}, allowing easy customization of individual components.
    In addition, SocialED provides several utility functions, including \texttt{utils.tokenize\_text} and \texttt{utils.construct\_graph} for data preprocessing, \texttt{metric} for evaluation metrics, and \texttt{utils.load\_data} for built-in datasets. 
    Leveraging popular deep learning libraries, SocialED simplifies detection tasks, ensuring efficient training and inference. 
    Its compatibility with standard data formats and pre-implemented algorithms enhances its flexibility and applicability across various event detection tasks. For more details, please refer to the SocialED documentation\footnote{\label{note2}Documentation: \url{https://socialed.readthedocs.io/}}.

\section{Library Robustness and Accessibility}

    \paragraph{Quality and Reliability}
    SocialED is built with robustness and high-quality standards, ensuring consistent performance. 
    As shown in Figure \ref{fig:vis}(c), we leverage continuous integration (CI) through \textit{GitHub Actions}\footnote{Continuous integration by GitHub Actions: \url{https://github.com/RingBDStack/SocialED/actions}} to automate testing across various Python versions and operating systems. 
    Unit tests are triggered automatically with every commit and pull request, and scheduled daily tests maintain ongoing stability. 
    The library achieves over 99\% code coverage, helping catch potential issues early. 
    SocialED is PyPI-compatible and adheres to PEP 625, making installation straightforward for modern Python environments. 
    The code follows the PEP 8 style guide, promoting readability and ease of collaboration.
    
    \paragraph{Accessibility and Community Support}
    SocialED is designed to be accessible to both beginners and experts. 
    The library offers detailed, user-friendly API documentation hosted on \texttt{Read the Docs}, with step-by-step guides, real-world examples, and tutorials to help users quickly understand and apply its features. 
    The API design, inspired by frameworks like \texttt{scikit-learn}, is intuitive and easy to integrate into existing workflows.
    For more advanced use cases, the documentation provides in-depth explanations, enabling users to adapt SocialED to their specific needs. 
    As an open-source project hosted on \texttt{GitHub}, SocialED encourages community engagement, with an easy-to-use issue-reporting mechanism and a clear contribution guide. 

\section{Conclusion and Future Plans}

    In this paper, we introduce SocialED, an open-source Python library that integrates state-of-the-art algorithm implementations and datasets for social event detection. 
    With its unified API, comprehensive documentation, and robust code design, SocialED serves as a valuable tool for both academic research and industrial applications.
    Looking ahead, the development of SocialED will focus on several key areas: 
    \textit{(1) Expanding Algorithms and Datasets}: Continuously integrating advanced algorithms and expanding datasets to cover a wider range of events across languages, fields, and cultures;
    \textit{(2) Enhancing Intelligent Functions}: Incorporating automated machine learning for model selection and hyperparameter optimization to improve algorithm performance and adaptability;
    and \textit{(3) Supporting Real-time Detection}: Enhancing real-time event detection and trend analysis capabilities, including support for streaming data to improve early warning and response.

\acks{We appreciate the hard work of all researchers whose open-source code contributed to this work. This work is supported by NSFC through grant 62322202.}

\bibliography{socialed}

\end{document}